\documentclass[runningheads]{llncs}
\usepackage{graphicx}
\usepackage{subfig}
\usepackage[english]{babel}
\usepackage[utf8]{inputenc}
\usepackage{amsmath}

\begin{document}
\title{Bayesian Volumetric Autoregressive generative models for better semisupervised learning}
\titlerunning{Bayesian Volumetric Pixel CNN}
\author{Guilherme Pombo \inst{1}, Robert Gray \inst{1}, Tom Varsavsky \inst{1, 2}, John Ashburner \inst{1}, Parashkev Nachev \inst{1}}
\authorrunning{G. Pombo}
\institute{Institute of Neurology, UCL \and School of Biomedical Engineering and Imaging Sciences, Kings College London, UK}
% \institute{Institute of Neurology, UCL}
% \institute{The Wellcome Trust Centre for Neuroimaging}
% \email{ucl@ac.uk}\\
% \url{ucl.ac.uk} \and
% UCL\\
% \email{\{rmapgcp,p.nachev,r.gray\}@ucl.ac.uk}}
%
\maketitle              % typeset the header of the contribution
\begin{abstract}
Deep generative models are rapidly gaining traction in medical imaging. Nonetheless, most generative architectures struggle to capture the underlying probability distributions of volumetric data, exhibit convergence problems, and offer no robust indices of model uncertainty. By comparison, the autoregressive generative model PixelCNN can be extended to volumetric data with relative ease, it readily attempts to learn the true underlying probability distribution and it still admits a Bayesian reformulation that provides a principled framework for reasoning about model uncertainty. \newline
Our contributions in this paper are two fold: first, we extend PixelCNN to work with volumetric brain magnetic resonance imaging data. Second, we show that reformulating this model to approximate a deep Gaussian process yields  a measure of uncertainty that improves the performance of semi-supervised learning, in particular classification performance in settings where the proportion of labelled data is low. We quantify this improvement across classification, regression, and semantic segmentation tasks, training and testing on clinical magnetic resonance brain imaging data comprising T1-weighted and diffusion-weighted sequences.

\end{abstract}
\section{Introduction}

There are two common problems with discriminative learning: class imbalance and sparse labels. These problems are particularly prevalent in medical imaging, due to the essential nature of clinical data. Semi-supervised learning provides a partial solution to these problems. Semi-supervised learning can be improved by using deep generative models, to learn better representations of the data, where generalisable decision boundaries are easier to identify \cite{semi_kingma}. \par
Variational autoencoders (VAEs), generative adversarial networks (GANs), and autoregressive (AR) models are the leading architectures for deep generative modelling. Unfortunately, their application to volumetric data has so far proved challenging, owing to poor convergence and distribution mode dropping, in the case of GANs \cite{glow}, or to potentially inaccurate error bounds and inappropriate independence assumptions, in the case of VAEs \cite{glow}. The use of generative modelling with high resolution 3D data is still only tentatively explored. \par

Our contributions are as follows: in \S3 we show how the 2D generative model PixelCNN \cite{pixel_rnn} can be extended to work efficiently with volumetric data. We call the resulting model 3DPixelCNN.  Furthermore, we incorporate the architectural changes suggested in \cite{dropout_bayes} so that we can compute voxel-wise measures of uncertainty with little computational overhead. In \S4 we show the benefits of using these uncertainty measures and 3DPixelCNN's hidden layer activations, in semi-supervised scenarios where labelled data is limited. Our evaluation incorporates three tasks: semantic segmentation of acute stroke lesions on diffusion weighted imaging (DWI) and age regression and sex classification on grey matter tissue compartments extracted from T1-weighted magnetic resonance imaging (MRI). Code available at https://github.com/guilherme-pombo/3DPixelCNN

%%%%%%%%%%%%%%%%%%%%%%%%%%%%%%%%%%%%%%%%%%%%%%%%%%%%%%%%%%%%%%%%%%%%%%
\section{Related work}
\subsection{Generative models for brain imaging}

We are interested in modelling \(p(\boldsymbol{x})\), the probability distribution for the stochastic process that generates our brain volumes. In the context of brain imaging, we have a likelihood model \(p_{\theta}\), where the parameters \(\theta\) are found by maximising the following objective:
\begin{equation}
    \mathcal{L}(\theta)=\frac{1}{N} \sum_{i=1}^{N} \log p_{\theta}\left(x_{i}\right) \sim \int p(\boldsymbol{x}) \log p_{\theta}(\boldsymbol{x}) d\boldsymbol{x};
    \label{likelihood}
\end{equation}
here, \(x_{1},...,x_{N}\) are the training volumes, which we assume have been sampled i.i.d from \(p(\boldsymbol{x})\). In medical imaging it is common to process volumes as 2D slices to reduce processing time and memory consumption. However, in order to utilise all of the information in \(x_{i}\), and to demonstrate the feasibility 3DPixelCNN, we use a fully 3D model. \par
To the best of our knowledge, \cite{nick_encoder} is the only work prior to ours to train a generative model on high-resolution 3D brain imagery. They model the (relatively low-detail) computed tomography (CT) modality using an approximation to a deep Gaussian process (c.f. \S\ref{deep_GP_section}) and an Autoencoder (AE). In the present article we also use this approximation but with a generative model that has increased representational power. We describe this model in the following section.

\subsection{PixelRNN}\label{prnn_sec}

In \cite{pixel_rnn}, the authors show how to model \(p(\boldsymbol{x})\) autoregressively, by modelling the joint distribution of pixels in an image using recurrent neural networks. They treat their (2D) images, with dimensions \(M \times N\), as a one-dimensional sequence of length \(MN\), and  they write the product of the conditional distributions over pixels as:

\begin{equation*}
    p(\boldsymbol{x})=\prod_{i=1}^{MN} p\left(x_{i} | x_{1}, \ldots, x_{i-1}\right).
\end{equation*}

This model is comparatively slow due to RNNs' difficulty in parallelising, so the authors approximate it with much faster standard convolutional networks. To ensure the receptive field of each convolution around each pixel only includes the pixels on which its probability is conditioned (thus, avoiding seeing the future context) they add masks to the convolutions. However, the bounded nature of this `masked' convolutional architecture causes a significant part of the input image to be ignored: a triangular pattern of omitted voxels they call the `blind spot'. To remedy this, the authors of \cite{c_pixel_cnn} use two masks instead of one, which they call `stacks': the first one is conditioned on the row so far (the `horizontal' stack) and the second one conditions on all rows above (the `vertical' stack). \par
The greater computational efficiency of PixelCNN compared with PixelRNN carries a cost in reconstruction quality. However, it has been shown \cite{c_pixel_cnn} that this can be ameliorated by replacing the rectified linear units between the convolutional stacks with a gated activation unit. This results in a better emulation of a long short-term memory (LSTM) gate. This use of both the convolutional stacks and the gated unit has enabled PixelCNN to match PixelRNN's reconstruction quality, whilst maintaining computational feasibility. \par

\subsection{Dropout as a Bayesian Approximation}\label{deep_GP_section}

Unlike VAEs, AR models are not Bayesian by construction, and they do not produce implicit or explicit estimates of model uncertainty. In \cite{dropout_bayes}, Gal and Ghahramani show that simply incorporating Dropout \cite{dropout} in every layer of any given neural network makes it capable of doing Bayesian inference, without harming performance. Once these changes are made, the standard deviation of a large-enough batch of forward passes yields a robust measure of uncertainty.  \par

In \cite{spatial_dropout} it is shown that since natural images exhibit strong spatial correlation, the feature map activations are strongly correlated - so applying standard Dropout to the kernels of the convolution operators is ill advised. Hence, they purpose a new dropout method, SpatialDropout, whereby for a given convolution feature tensor of size \( \text{H} \times \text{W} \times \text{D} \times \text{channels} \), a mask of size \(1 \times 1 \times 1 \times \text{channels} \) is applied. \newline

%%%%%%%%%%%%%%%%%%%%%%%%%%%%%%%%%%%%%%%%%%%%%%%%%%%%%%%%%%%%%%%%%%%%%%
\section{Methods}

To extend the PixelCNN solution to volumetric data we must first solve the blind spot problem for 3D (c.f. \S \ref{prnn_sec}). Consider our model processing an \(M \times N \times K\) volume, and currently calculating the conditional distribution of the voxel with coordinates \((R,C,D)\), which we denote \(x_{R, C, D}\). We must now use three stacks (c.f. \S \ref{prnn_sec}): horizontal, depth and vertical. \par
The \textbf{Horizontal stack} conditions on the current depth channel and takes as input the output of the previous horizontal stack gate, as well as the output of the depth and vertical stacks. The set of voxels it considers is \(\{x_{R, C, d} | d \in \{1, \ldots, D-1 \} \}\). In turn the \textbf{Depth stack} conditions on all the entries to the left of the current voxel, but does not go up any rows. It takes as input the output of the previous depth gate, as well as the output of the vertical stack. Its receptive field grows in 2D rectangular fashion, defined by the set \(\{x_{R, c, d} | c \in \{1, \ldots, C-1 \}, d \in \{1, \ldots, K \}\}\).  Finally, the \textbf{Vertical Stack}  conditions on all the rows and columns in the level above the current voxel. It does not have any masking. Its output is fed into the horizontal and depth stacks and its receptive field grows as a cuboid, defined by the set \(\{x_{r, c, d} | r \in \{1, \ldots, R-1\}, c \in \{1, \ldots, N\}, d \in \{1, \ldots, K \} \}\). \par
\begin{figure}
    \centering
    \includegraphics[width=8cm]{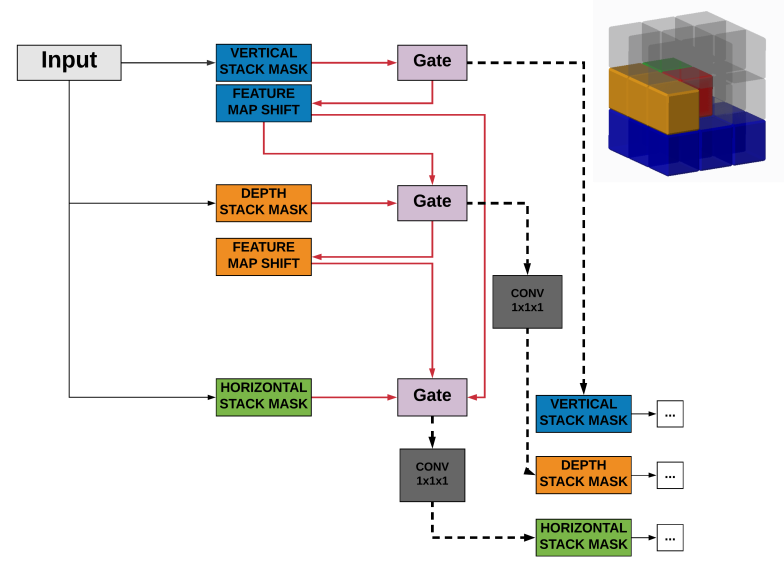}
    \caption{The two figures show how the vertical (blue), depth (orange) and horizontal(green) stacks, are used to get the conditional distributions over the pixels, for the pixel in consideration (red) (Cube upside down for easier visualisation)}
    \label{fig:masked_convs}
\end{figure}

These stacks ensure our convolution operations have the correct receptive fields. To reiterate, using just regular convolutions would lead to a bounded receptive field, which in turn would have led to the omission of several voxels from calculations of the conditional distribution (a pyramidal `blind spot'). These stacks are represented in figure \ref{fig:masked_convs}. We use the gated activation unit from \cite{c_pixel_cnn} to efficiently combine the information of different stacks. We first add the stacks together and do a channel-wise split. If the tensor has \(N\) channels, then we now have tensor \(W_1\) with the first \(N/2\) channels and tensor \(W_2\) with the remaining channels. The gated activation unit is calculated as \(\tanh (W_1) \odot sigmoid(W_2)\), where \(\odot\) is the Hadamard product.
After each gate we have a skip shortcut \cite{skip_connections} to the next stack in the model. After the first layer, as in \cite{pixel_cnn_pp} we also add a residual connection \cite{skip_connections} from a Gated unit to the next one. SpatialDropout is applied after every convolution operator so that we can approximate a deep Gaussian process (see \S2.2). Model statistics are derived at test time from batches of multiple forward passes with dropout enabled. We denote the mean and standard deviation of these batches by \(\boldsymbol{\mu}\) and \(\boldsymbol{\sigma}\) respectively. \par

We train our 3DPixelCNN models using continuous negative log likelihood (NLL), and evaluate using log likelihood. We used continuous rather than discrete NLL as it has been shown \cite{pixel_cnn_pp} that treating pixel intensities as emission probabilities performs poorly for large images, resulting in noisy and speckly reconstructions. We trained for 20 epochs using the Adam optimiser \cite{adam}. The initial learning rate was 0.001, the batch size was 1 and the dropout rate was 0.15 (dropout rates between 0.1 and 0.2 are recommended in \cite{dropout_bayes}). Our model has five layers with the structure depicted in figure \ref{fig:masked_convs}. We use kernel sizes of \(3\times3\times3\) for all non-masked convolutions in the network. We could have incorporated downsampling as in \cite{pixel_cnn_pp}, but we leave this for future work. \par

\section{Experiments and results}

\textbf{Data}: We use two separate datasets. One is a collection of routinely acquired DWI from patients evaluated for acute stroke at our clinic. This comprises 1333 scans with evidence of an acute ischaemic lesion, and 982 scans with no evidence of an acute lesion but variable presence of chronic vascular disease. The volumes we use consist of the b1000 sequence non-linearly registered to MNI space with unified segmentation \cite{john}. A manually-curated binary mask delineating the area of ischaemic damage is our ground truth for lesion semantic segmentation \cite{tianbo}. We also use a manually curated mask to remove any voxels outside of the brain.  \par

The second dataset consists of 13287 SPM Grey Matter (GM) tissue compartments from MRIs obtained from UK Biobank, and routinely acquired clinical imaging at UCLH. The GM segmentations were derived using methods from \cite{john}. Sex and age are known for all patients and were used to evaluate models on classification and regression tasks. For both modalities we reduced the computational burden (due to time constraints) by downsampling the volumes, using bilinear resampling, to 3mm resolution \(52\times64\times52\) volumes. \par

\textbf{Image reconstructions}: For each volume in the DWI and GM datasets, we produce its reconstruction, and then generate \(\boldsymbol{\mu}\) and \(\boldsymbol{\sigma}\) by performing \(T=20\) forward passes with dropout left on (c.f. \S \ref{deep_GP_section}). \par
We use a train/validation/test split of 80/10/10. The best log likelihood obtained by the model in the task of volume reconstruction on the test sets at 3mm, are \textbf{0.360} for the DWI data and \textbf{0.105} for the GM data. Our model outperforms the Bayesian AE from \cite{nick_encoder} which achieves 0.378 on DWI and 0.222 on GM. Notice that on the more detailed modality (T1-GM) our model performs 111\% better.\par

\begin{figure}%
    \centering
    \subfloat[DWI bayesian reconstructions]{{\includegraphics[width=6.2cm, height=5cm]{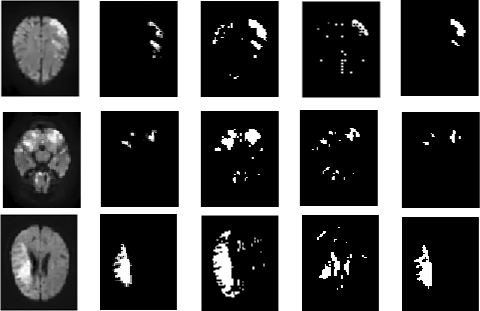} }}%
    \qquad
    \subfloat[MRI bayesian reconstructions]{{\includegraphics[width=5cm]{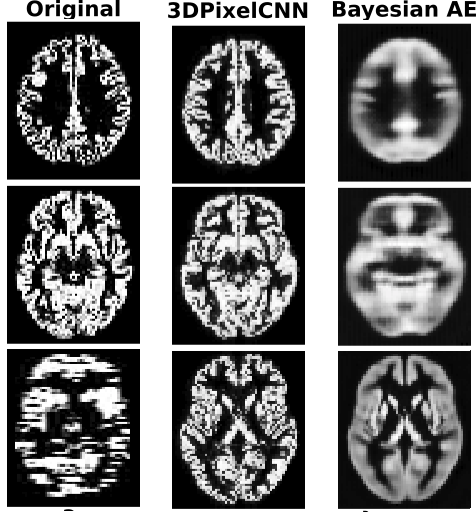} }}%
    \caption{\textbf{(a)}: From left to right: 1) The slice through the axial plane with the greatest area of lesion, 2) The stroke label map , 3)  \(\tau(x_i)\) 4) \(|x_i - AE(x_i)|\) 5) \(\tau(\boldsymbol{\sigma})\). \(\boldsymbol{\sigma}\) helps capture the tighest bound on the lesion \textbf{(b)} Axial slices of 1) The original volume, 2) The 3DPixelCNN reconstruction and 3) The Bayesian AE reconstruction (On the last volume there was a capture problem and we use it to test 3DPixelCNN's ability to super resolve) }%
    \label{fig:recons}%
\end{figure}

In order to produce uncertainty estimates (\(\boldsymbol{\sigma}\)) for DWI, we trained our 3DPixelCNN only on data with no evidence of stroke lesion, i.e. from the distribution \(p(\boldsymbol{x}|\text{no lesion})\). Therefore, when producing \(\boldsymbol{\sigma}\) for lesioned data, the uncertainty masks provide a measure of the distance from the lesioned brain to the expected distribution of non-lesioned brains. We use a simple classification strategy on the volumes -  thresholding the average intensity of the volume, \(x_i\), which we denote as \(\tau(x_i)\). On the DWI ischemic stroke lesion test set, applying this classification strategy on regular volumes yields Dice coefficients of 14.7 \%, whereas on \(\boldsymbol{\sigma}\) it yields Dice coefficients of 23.7 \%. This same strategy on the Bayesian AE, \(|x_i - AE(x_i)|\) (see \cite{nick_encoder} for more details)  yields a performance of 17.3 \%. This provides early confirmation that uncertainty estimates of generative models capture useful task-independent signal. \newline
Figure \ref{fig:recons} shows a representative selection of reconstructions of GM volumes and unsupervised lesion masks produced using \(\tau(x_i)\). Notice on the MRI reconstruction, when the original image is corrupted, the 3DPixelCNN model acts as a super resolution mechanism, further showing the model has learnt \(p(\boldsymbol{x})\) and is not simply memorising the training set.

\textbf{Semi-supervised learning}: To experiment with using our uncertainty measures to improve supervised tasks, we use our DWI dataset for evaluating models on the task of \textbf{semantic segmentation} and our GM dataset to evaluate \textbf{regression} and \textbf{classification} tasks. \par
For the segmentation task we use a 3D U-Net \cite{unet} as the \textbf{baseline}. As the DWI dataset is not yet public, there are no state of the art results to which we can compare our results. For the age regression and sex classification tasks we use the architecture from \cite{age} as our baseline, which we'll call ASC, adding only L2 regularisation and Dropout to ensure better generalisation. All models are trained with early stopping using the validation set, the criterion being 20 successive epochs without a drop in validation error. The models are trained in 5-fold fashion (80/10/10 split) for added statistical resilience. We compare the models' Dice scores on a semantic segmentation task, their mean absolute errors on an age regression task and their binary accuracy on a sex classification task, all evaluated on the test set. \par

\begin{figure}
    \centering
    \includegraphics[width=11cm]{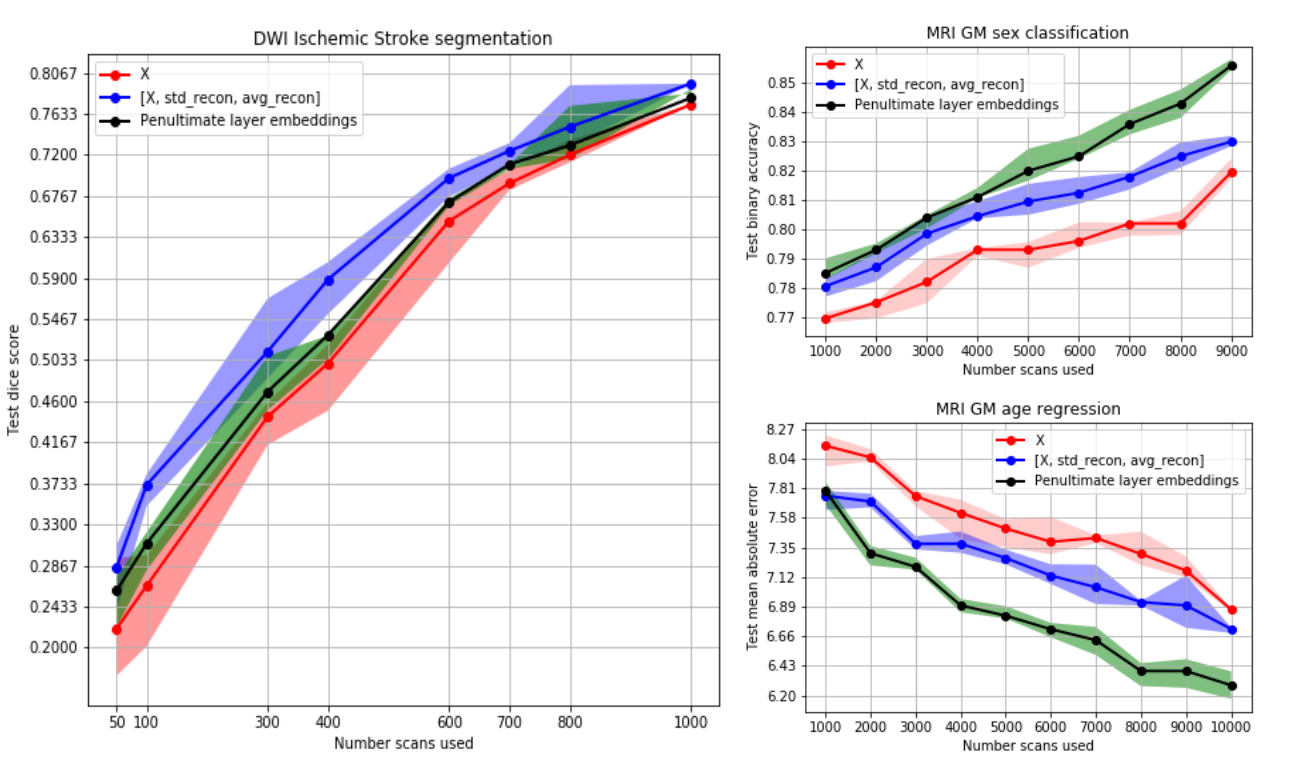}
    \caption{From left to right: Comparison of DWI segmentation performance, Comparison of GM sex classification performance and age regression performance}
    \label{fig:evals}
\end{figure}

Figure \ref{fig:evals} shows mean model performance with error bars for three different types of inputs into both the 3D-UNet and ASC classifiers: (1) using just the original volumes as input (red- \(\boldsymbol{\chi}\)); (2) using original data concatenated with \(\boldsymbol{\mu}\) and \(\boldsymbol{\sigma}\) (blue - \(\boldsymbol{\xi}\)). For the case Bayesian AE we concatenate \(\boldsymbol{\mu}\) and \(|x_i - AE(x_i)|\); (3) using the activations of the penultimate convolutional layer of the 3DPixelCNN. (black/green - \(\boldsymbol{\psi}\)). For the Bayesian AE we use its latent space. \par
When using 3DPixelCNN, we notice that performance with \(\boldsymbol{\xi}\) was significantly better than with \(\boldsymbol{\chi}\), for all dataset sizes tested and classification tasks. For sex classification and age regression, using \(\boldsymbol{\psi}\) results in better performance than both \(\boldsymbol{\chi}\) and \(\boldsymbol{\xi}\). We speculate that this is because the embeddings, which are higher-dimensional (10 vs 3 channels), comprise a decomposition of the data from which useful decision boundaries can be more readily identified, although this extra dimensionality comes at the cost of greater GPU memory requirements. On the other hand, for lesion segmentation, using \(\boldsymbol{\xi}\) performs better than using either \(\boldsymbol{\chi}\) or \(\boldsymbol{\psi}\). \par
For semantic segmentation using \(\boldsymbol{\xi}\), the increase is most noticeable at smaller \(N\) with an improvement of 0.082 (25.6\%) in Dice coefficient for \(N < 500\) and an average increase of 0.056 (15.2\%) for all \(N\). Using \(\boldsymbol{\psi}\) provides less of a performance gain, with an average increase in Dice of 0.025 (6.9\%). For age regression and sex classification, we notice a steady increase in performance when using \(\boldsymbol{\xi}\), with an average error reduction of 0.30 years (3.98\%) and accuracy increase of 1.87\%, respectively. Using \(\boldsymbol{\psi}\), on the other hand, results in an average error reduction of 0.68 years (9.09\%) for age regression and accuracy increase of 3.36\%, for sex classification. Using the Bayesian AE's \(\boldsymbol{\xi}\) results in a performance degradation of at least 2\% for all tasks, compared to using the original volume. We suspect this is because here \(\boldsymbol{\xi}\) is relatively noisy, as can be seen in Fig. \ref{fig:recons}. On the other hand, using the latent space, \(\boldsymbol{\psi}\), results in an average 5.6\% increase for the age regression task and a 2.2\% increase for sex classification. The Bayesian AE's latent space degraded performance for the semantic segmentation task. \par

Clearly, 3DPixelCNN's uncertainty measures help most with semantic segmentation. They seem to be most useful for tasks with more localised signal (lesion segmentation) as opposed to global signal. We speculate this is  because in the lesioned brains \(\boldsymbol{\sigma}\) is more focused on the lesion, since we had the generative model learn \(p(\boldsymbol{x}|\text{no lesion})\), whereas the uncertainty maps are much noisier for volumes with less obvious abnormalities, since the 3DPixelCNN learnt only \(p(\boldsymbol{x})\). We hypothesize these uncertainty measures are also helpful in the presence of artifacts (as can be seen in Figure  \ref{fig:recons}), which is why they also helped for tasks with less abnormal brains. 

\section{Conclusion}

We have presented the first implementation of a volumetric neural network-based autoregressive model. We have shown that it is a method that can capture the richness of a complicated 3D probability distribution and is therefore well-suited to medical imaging. By augmenting labelled data with measures of uncertainty derived from unsupervised models, we saw improved performance in every supervised task we carried out. For tasks on brains without gross abnormalities, we found it was better to use 3DPixelCNN's penultimate layer activations than the uncertainty estimates. For lesion detection, we found that the uncertainty measures provided a bigger performance increase, which is of more utility in the medical imaging domain. \textbf{Acknowledgments}: This research has been conducted using the UK Biobank Resource under Application Number 16273. This work is supported by the EPSRC-funded UCL CDT in Medical Imaging (EP/L016478/1), the Department of Health’s NIHR-funded BRC at UCLH and the Wellcome Trust.
%\bibliographystyle{splncs04}
%\bibliography{paper_479_bibliography}

\end{document}